\documentclass[conference]{IEEEtran}
\IEEEoverridecommandlockouts
\usepackage{amsmath,amssymb,amsfonts}
\usepackage{algorithmic}
\usepackage{graphicx}
\usepackage{textcomp}
\usepackage{xcolor}
\usepackage{url}
\usepackage{wrapfig}
\usepackage{float}
\usepackage{placeins}
\usepackage{titlesec}
\usepackage{enumitem}

\usepackage[style=numeric,sorting=none
]{biblatex}
\AtBeginBibliography{\footnotesize}

\usepackage[font=footnotesize,skip=4pt]{caption}
\def\BibTeX{{\rm B\kern-.05em{\sc i\kern-.025em b}\kern-.08em
    T\kern-.1667em\lower.7ex\hbox{E}\kern-.125emX}}
\begin{document}

\title{Enhancing Road Safety Through Multi-Camera Image Segmentation with Post-Encroachment Time Analysis\\
}

\author{\IEEEauthorblockN{Shounak Ray Chaudhuri}
\IEEEauthorblockA{\textit{Department of Electrical and}\\
\textit{Computer Engineering} \\
\textit{San Diego State University}\\
\textit{California, USA}\\
ORCID: 0009-0003-8084-7104}
\and
\IEEEauthorblockN{Arash Jahangiri}
\IEEEauthorblockA{\textit{Department of Civil, Construction,}\\
\textit{and Environmental Engineering} \\
\textit{San Diego State University}\\
\textit{California, USA}\\
ORCID: 0000-0002-8825-961X}
\and
\IEEEauthorblockN{Christopher Paolini}
\IEEEauthorblockA{\textit{Department of Electrical and}\\
\textit{Computer Engineering} \\
\textit{San Diego State University}\\
\textit{California, USA}\\
ORCID: 0000-0001-6563-917X}
}

\maketitle

\begin{abstract}
Traffic safety analysis at signalized intersections is essential for reducing vehicle and pedestrian collisions, yet traditional crash-based studies are limited by data sparsity and reporting latency. This paper presents a multi-camera computer vision framework for real-time safety assessment through Post-Encroachment Time (PET) computation, demonstrated at the intersection of H Street and Broadway in Chula Vista, California. Four synchronized cameras provide continuous visual coverage, with frames processed on NVIDIA Jetson AGX Xavier edge devices using YOLOv11 segmentation for vehicle detection. Detected vehicle polygons are transformed into a unified bird’s-eye map via homography, enabling alignment across overlapping camera views. A pixel-level PET algorithm tracks temporal occupancy at each spatial location by measuring the time between successive vehicle passages, enabling fine-grained hazard visualization through dynamic heatmaps with 3.3 sq-cm spatial resolution. Timestamped vehicle trajectories and PET data are stored in an SQL database for longitudinal analysis. Results across multiple time intervals demonstrate the framework’s ability to identify high-risk regions with sub-second temporal sensitivity and real-time edge throughput, generating 800 × 800 logarithmic heatmaps at an average of 2.68 FPS. This paper validates decentralized vision-based PET analysis for intelligent transportation systems and presents a scalable methodology for high-resolution, real-time intersection safety evaluation.
\newline
\end{abstract}

\begin{IEEEkeywords}
Post-Encroachment Time (PET), Traffic Safety Analysis, Vehicle Detection, Intelligent Transportation Systems (ITS), Computer Vision, Multi-Camera Systems, Homography, Bird's-eye Mapping, Intersection Monitoring, YOLOv11, Edge Computing, NVIDIA Jetson, Traffic Conflict Analysis, Image Segmentation, Real-Time Processing
\end{IEEEkeywords}

\section{Introduction}

\subsection{Context and Motivation}

Traffic safety at signalized intersections remains a critical concern in urban planning, as intersections involve high vehicle conflict and elevated accident risk. Large intersections pose particular challenges due to increased maneuvering space, multiple conflict points, and reduced natural traffic control, which can lead to higher speeds and greater uncertainty in driver behavior.

As traffic volume rises, the likelihood of collisions grows nonlinearly, exacerbated by limited gaps between vehicles and dense lane usage. Quantifying these risks is essential for designing interventions, optimizing traffic control, and enabling data-driven safety assessments that help mitigate accidents and improve overall traffic efficiency.

\subsection{Problem Statement}

The intersection of H Street and Broadway in Chula Vista, California, is a busy urban junction with high traffic volumes, multiple signal phases, and pedestrian activity, increasing collision risk. To enable data-driven safety analysis, San Diego State University researchers partnered with the City of Chula Vista to install four pan-tilt-zoom (PTZ) traffic cameras on the intersection’s signal poles. A local Ethernet network in a traffic cabinet with four NVIDIA Jetson AGX Xavier devices facilitated continuous video capture and processing for evaluating traffic behavior and collision risk.

The site’s history of congestion and injurious traffic incidents, including pedestrian-involved crashes, makes it ideal for computing quantitative traffic safety metrics, such as Post-Encroachment Time (PET). PET measures the temporal gap between vehicles or between vehicles and pedestrians at a specific location, identifying potential conflicts. In this project, a pixel-level PET approach was used: each pixel acts like a “stopwatch” that counts up while unoccupied and resets when a vehicle passes, with the maximum observed values recorded over time. This allows fine-grained analysis of how long areas remain clear and precise identification of high-risk locations.

\subsection{Contributions}

This paper presents the following contributions:

\begin{itemize}[leftmargin=*, labelsep=0.5em]
\item \textbf{High-resolution, multi-camera dataset for PET analysis:} Four synchronized cameras provided timestamped vehicle trajectory data across all lanes and approaches.
\item \textbf{Pixel-level PET computation:} A stopwatch-based framework tracks occupancy at each pixel over time, capturing the duration between vehicle visits for precise detection of close-encounter events beyond pre-defined cell boundaries.
\item \textbf{Camera overlap and homography integration:} Multiple camera views were combined using homography matrices to align overlapping fields of view, improving vehicle localization and enabling 3D insights from 2D video.
\item \textbf{Real-world validation and wide deployability:} Techniques were validated in a real urban intersection and are broadly applicable to similarly complex intersections worldwide.
\item \textbf{Fine-grained hazard visualization:} PET values accurate to 3.3 × 3.3 sq-cm pixels were expressed as continuous heatmaps, revealing specific high-risk areas for vehicle-to-vehicle or vehicle-to-pedestrian conflicts.
\item \textbf{Actionable safety insights:} Long-term analysis highlights critical zones and supports evaluation of architectural or policy interventions to reduce collision probability at the studied intersection.
\end{itemize}

\section{Related Work}

\subsection{Vehicle Detection}

Vehicle detection is fundamental in intelligent transportation systems (ITS), autonomous driving, traffic monitoring, and urban planning, enabling downstream applications such as traffic flow analysis, collision avoidance, and automated toll or parking systems. Early methods relied on hand-crafted features like Haar cascades, Histogram of Oriented Gradients (HOG), and deformable part models (DPM) to extract vehicle shapes~\cite{DOC_SOURCE_29, DOC_SOURCE_30, DOC_SOURCE_31}, but these approaches struggled under varying lighting, occlusions, and complex urban scenes.

Deep learning, especially convolutional neural networks (CNNs), has largely replaced classical methods. CNN-based detectors are broadly categorized as multi-stage (e.g., Faster R-CNN) and single-stage (e.g., YOLO series, SSD) detectors~\cite{DOC_SOURCE_34, DOC_SOURCE_32, DOC_SOURCE_33}, with single-stage detectors favored for real-time applications~\cite{DOC_SOURCE_7, DOC_SOURCE_8}. YOLO (You Only Look Once) treats detection as a regression problem, predicting bounding boxes and class probabilities simultaneously~\cite{DOC_SOURCE_32}. A recent iteration, YOLOv11, improves performance under occlusions and varying illumination using enhanced backbones, anchor-free detection, and attention modules~\cite{DOC_SOURCE_9, DOC_SOURCE_10}. Other approaches, including R-CNN variants, RetinaNet, and transformer-based methods, have further advanced detection in crowded or unstructured environments and are increasingly combined with trajectory prediction and behavior analysis~\cite{DOC_SOURCE_11, DOC_SOURCE_12, DOC_SOURCE_13}.

\subsection{Traffic Safety Metrics}

Post-Encroachment Time (PET) is a widely used surrogate measure of traffic conflict at intersections, defined as the time between a leading vehicle leaving a conflict point and a following vehicle arriving~\cite{DOC_SOURCE_14, DOC_SOURCE_15}. Lower PET indicates higher collision risk, providing a useful proxy where crash data are sparse~\cite{DOC_SOURCE_1}.

PET has been applied to left-turn maneuvers~\cite{DOC_SOURCE_14, DOC_SOURCE_15}, integrated into automated safety evaluation frameworks~\cite{DOC_SOURCE_1}, and combined with measures like extended delta-v to assess micro-level interactions~\cite{DOC_SOURCE_17, DOC_SOURCE_18}. Advances in video analysis and sensors enable automated PET computation: right-turn conflicts have been evaluated via video~\cite{DOC_SOURCE_19}, large-scale multi-camera frameworks developed~\cite{DOC_SOURCE_20}, and deep learning applied to pedestrian-vehicle interactions~\cite{DOC_SOURCE_21}. These methods improve the accuracy and spatial resolution of PET for real-time traffic safety assessment.

\subsection{Bird's-eye Mapping}

Bird's-eye mapping, or top-down projection, transforms camera images into orthographic representations to simplify spatial reasoning and vehicle tracking, facilitating computation of metrics like PET or Time-to-Collision~\cite{DOC_SOURCE_12, DOC_SOURCE_19}. Homography matrices map image points to global coordinates, integrating overlapping cameras into unified views~\cite{DOC_SOURCE_12, DOC_SOURCE_19, DOC_SOURCE_22, DOC_SOURCE_20}. Recent work refines this for complex intersections, occlusions, and dynamic traffic scenes, incorporating calibration adjustments, distortion correction, and probabilistic localization~\cite{DOC_SOURCE_19, DOC_SOURCE_13}. Bird's-eye mapping underpins modern traffic safety analyses, enabling surrogate metric computation and visualization of vehicle interactions.

\begin{figure}[ht]
\centering
\includegraphics[width=0.8\linewidth]{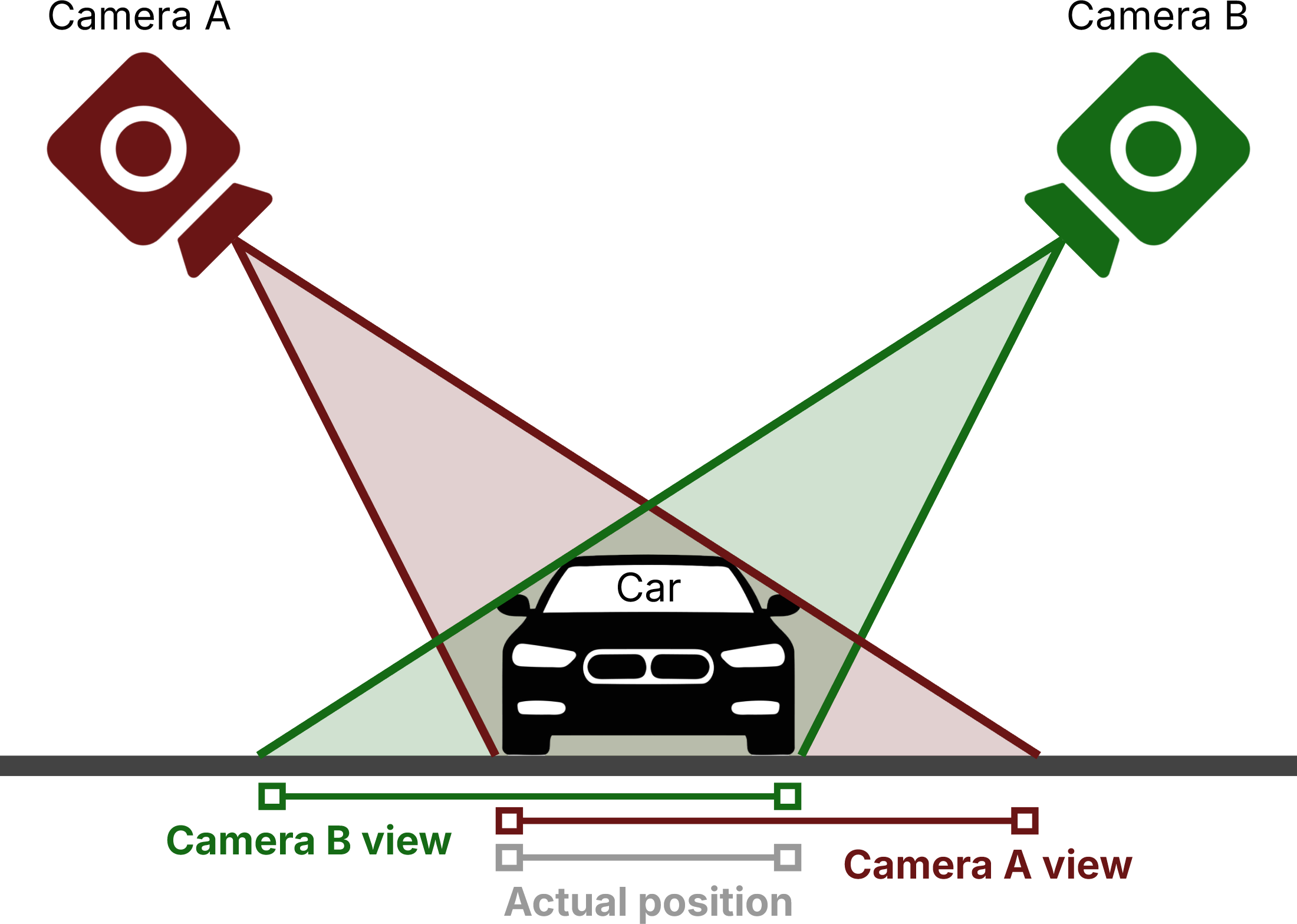}
\caption{Predicted vehicle locations from two overhead cameras.}
\label{fig:car_camera_overlap}
\end{figure}

Using homography matrices inherently assumes that all detected objects lie at ground level. However, as illustrated in Fig.~\ref{fig:car_camera_overlap}, this assumption can introduce inaccuracies: regions of the intersection occluded from an overhead camera’s perspective are treated as occupied, even if no vehicle is physically present in that space. To mitigate this, multiple cameras, such as cameras A and B in Fig.~\ref{fig:car_camera_overlap}, are employed. By combining overlapping views, the occlusion effects of individual cameras are canceled out, allowing the dataset to accurately capture the true positions of vehicles.

\subsection{Urban Traffic Studies}

Urban traffic studies examine vehicle interactions, intersection safety, and high-density road networks. Signalized intersections are particularly challenging due to multiple lanes, turning movements, and vulnerable road users~\cite{DOC_SOURCE_14, DOC_SOURCE_1}. Microscopic and macroscopic simulations model traffic behavior and safety indicators, while trajectory data capture real-world movements for metrics like Post-Encroachment Time (PET), Time-to-Collision (TTC), and conflict severity~\cite{DOC_SOURCE_24, DOC_SOURCE_25, DOC_SOURCE_26, DOC_SOURCE_13}. High-resolution, spatiotemporal datasets and connected vehicle data further enhance risk assessment and near-miss prediction~\cite{DOC_SOURCE_27, DOC_SOURCE_28}. Multi-camera and sensor-based approaches enable automated traffic monitoring, integrating detection, tracking, and bird's-eye mapping to identify critical zones and evaluate interventions for intersection safety.

\section{Methods}

\subsection{Data Acquisition}

Video data were collected at the intersection of H Street and Broadway in Chula Vista, California, using four fixed overhead Hikvision pan-tilt-zoom (PTZ) cameras mounted adjacent to the traffic lights. Each camera was positioned to maximize coverage of the intersection. The cameras were connected to an Ethernet switch forming a local area network (LAN). Four NVIDIA Jetson AGX Xavier computing devices and one Windows system were also connected to the LAN. The network included a Verizon LTE router to provide Internet access.

\begin{figure}[H]
    \centering
    \includegraphics[width=1\linewidth]{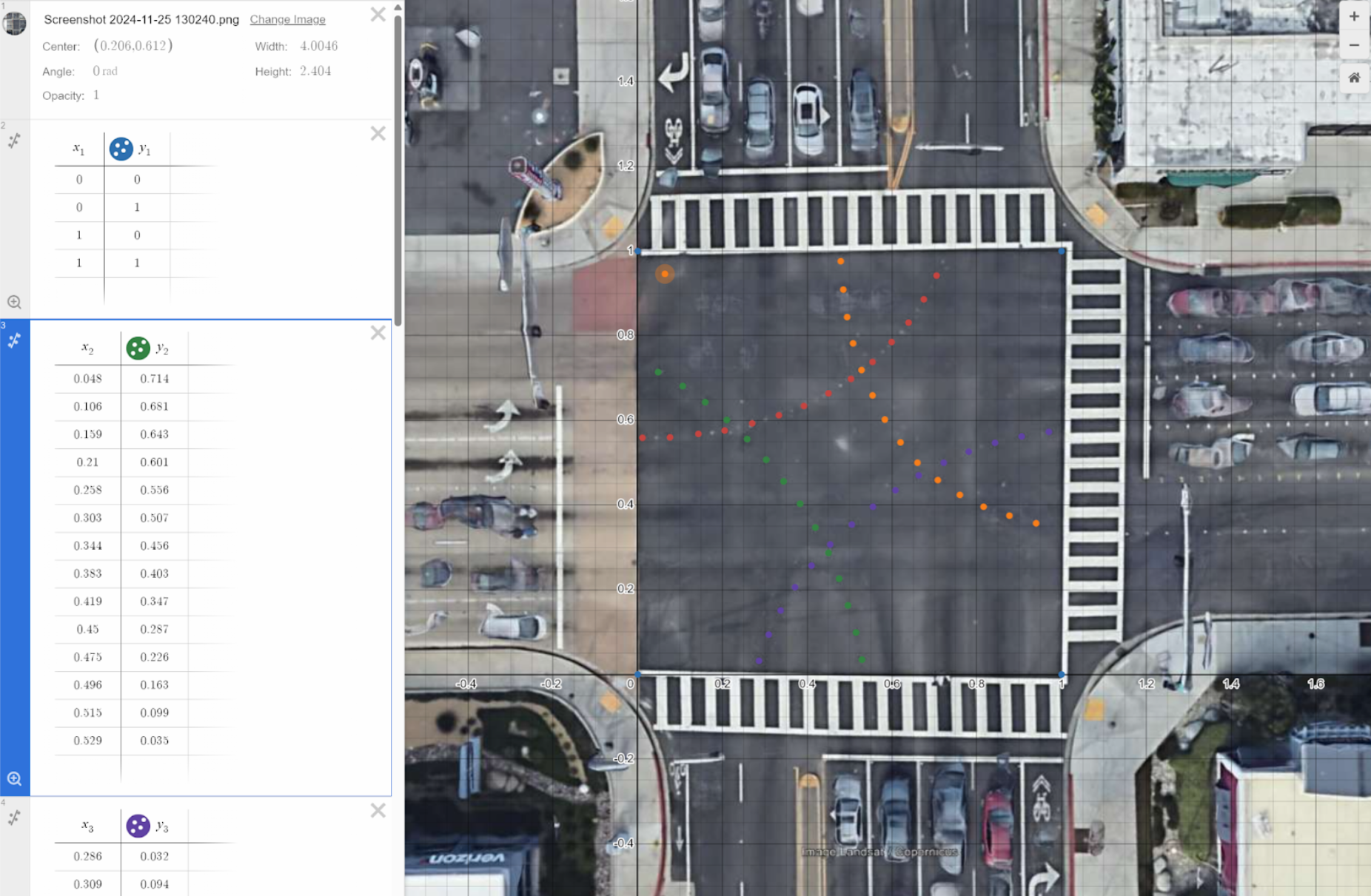}
    \caption{Intersection of H St and Broadway in Chula Vista, California, with annotated points of interest where annotations are Cartesian pixel values in global coordinates.}
    \label{fig:intersection}
\end{figure}
\FloatBarrier

\begin{figure}[ht]
    \centering
    \includegraphics[width=0.9\linewidth]{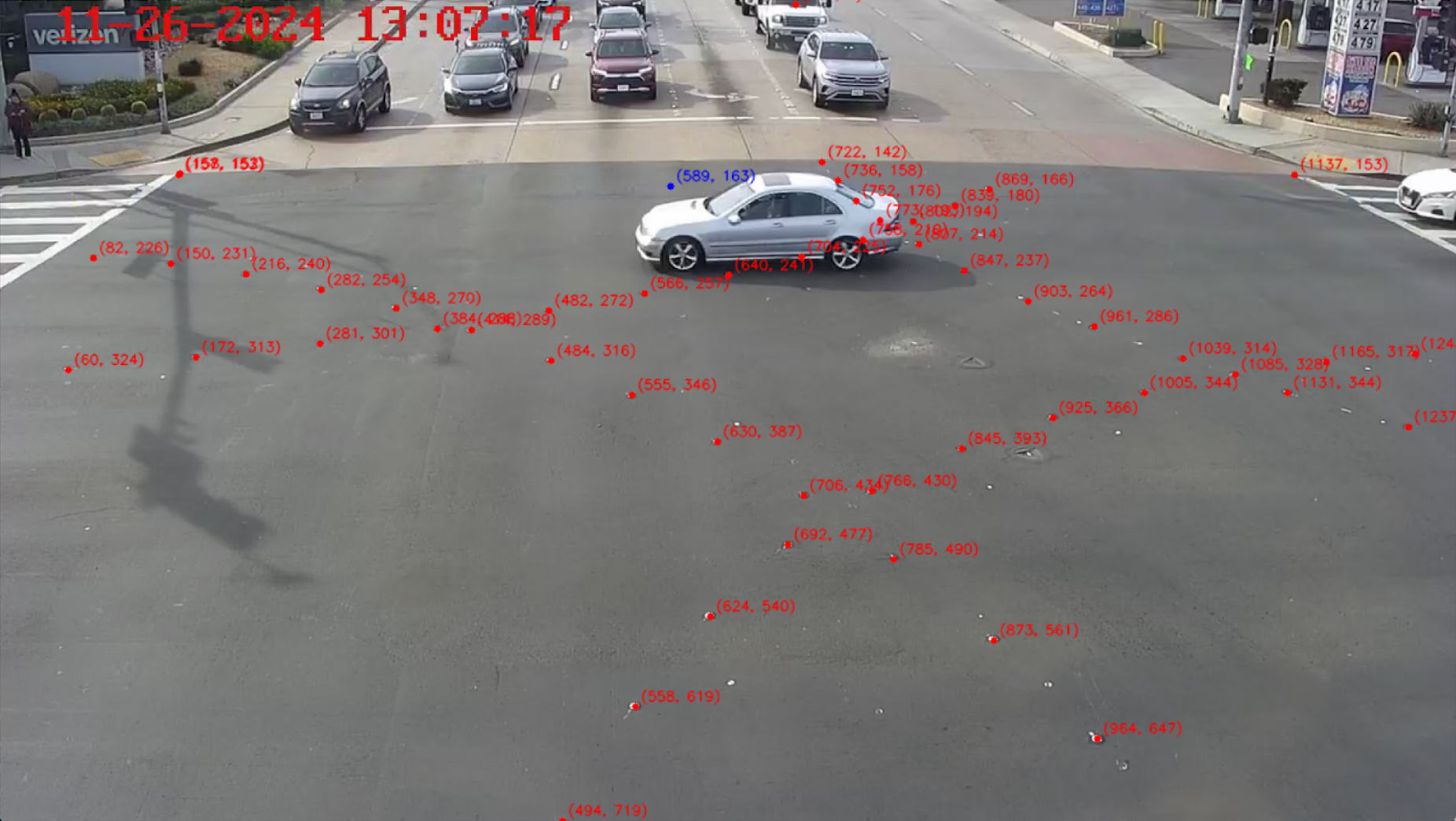}
    \caption{Camera perspective view of H Street and Broadway in Chula Vista, CA, with annotated points of interest.}
    \label{fig:annotated_points}
\end{figure}

The intersection was also identified on Google Earth. Coordinates of points of interest were recorded in both global coordinates (via Google Earth) and camera coordinates, as shown in Fig.~\ref{fig:intersection} and Fig.~\ref{fig:annotated_points}. Table~\ref{tab:global_camera_coords} illustrates the data storage format, showing five of the 31 total coordinates visible across all four cameras.

The video streams were accessed via RTSP for computation. After each camera was manually adjusted, a three-hour sample of timestamped video from all four cameras was recorded and uploaded to an off-site computer. This dataset was used exclusively for model development and testing, while subsequent computations were executed directly on the local NVIDIA Jetson devices.

\begin{table}[ht]
\caption{Corresponding points in global and camera coordinates}
\centering
\scriptsize
\resizebox{1\linewidth}{!}{%
\begin{tabular}{|cc|cc|cc|cc|cc|}
\hline
\multicolumn{2}{|c|}{\textbf{Global}} & \multicolumn{2}{c|}{\textbf{8005}} & \multicolumn{2}{c|}{\textbf{8006}} & \multicolumn{2}{c|}{\textbf{8007}} & \multicolumn{2}{c|}{\textbf{8008}} \\
\hline
\textbf{X} & \textbf{Y} & \textbf{X} & \textbf{Y} & \textbf{X} & \textbf{Y} & \textbf{X} & \textbf{Y} & \textbf{X} & \textbf{Y} \\
\hline
0.270 & 0.593 & 813 & 75  & 740 & 421 & 373 & 189 & 1232 & 267 \\
0.333 & 0.612 & 854 & 99  & 671 & 348 & 482 & 170 & 1142 & 283 \\
0.392 & 0.634 & 905 & 123 & 607 & 292 & 577 & 149 & 1057 & 302 \\
0.450 & 0.664 & 968 & 150 & 542 & 246 & 662 & 129 & 968  & 331 \\
0.503 & 0.698 & 1046 & 179 & 475 & 210 & 733 & 108 & 878  & 368 \\
\hline
\end{tabular}
}
\label{tab:global_camera_coords}
\end{table}

\subsection{Vehicle Detection}

YOLOv11, a single-stage object detection architecture, was employed in this project for real-time vehicle detection across multiple camera feeds~\cite{DOC_SOURCE_9}. This model integrates advanced feature extraction within a unified framework, enabling high-speed, high-accuracy vehicle detection. The medium-size segmentation variant, \textit{yolov11m-seg}, was utilized to output shaded regions corresponding to detected vehicles. Vehicle locations were computed, timestamped, and recorded simultaneously for all four camera streams.

\subsection{Polygon Coordinate Conversion}

To consolidate vehicle detections from the four cameras into a single top-down view, a homography-based projection method was applied. For each camera, a 3 $\times$ 3 homography matrix $\mathbf{H}$ was computed using OpenCV’s \textit{cv2.findHomography()} function with previously identified points of interest as reference.

\begin{figure}[ht]
    \centering
    \includegraphics[width=0.9\linewidth]{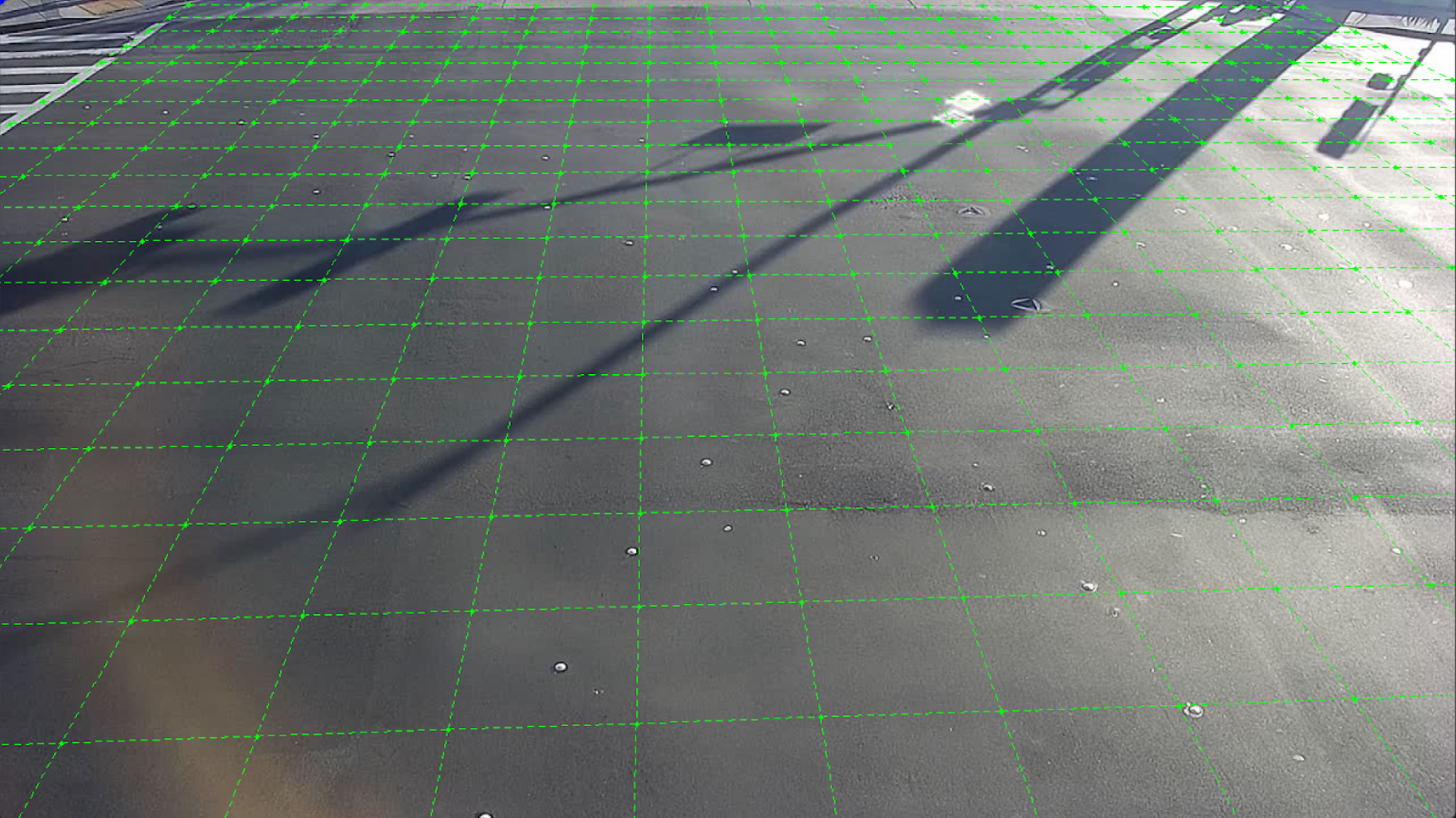}
    \caption{Visual result of applying the homography matrix of one of the cameras to convert coordinates.}
    \label{fig:overlaid_grid}
\end{figure}

Each detected vehicle polygon was projected into the global coordinate system by multiplying the polygon coordinates by $\mathbf{H}$ and performing homogeneous normalization. The resulting coordinates were scaled to a fixed-resolution grid to generate a unified bird’s-eye occupancy map of the intersection, as illustrated in Fig.~\ref{fig:overlaid_grid}.

The coordinate-converted polygons were serialized as JavaScript Object Notation (JSON) files, timestamped, and transmitted to the Windows system for subsequent rectangle fitting.

\subsection{Rectangle Fitting}

Coordinate-converted polygons were projected onto a unified world grid and converted into binary masks. These masks were summed to produce an overlap count matrix representing the number of cameras reporting a vehicle for each pixel. Point values were assigned to pixels based on the number of cameras detecting a vehicle. After manual experimentation, optimal point values for pixels reported by one, two, three, and four cameras were determined to be 1, 2, 6, and 8, respectively. An overlap mask highlighting areas with high agreement (three or four cameras) was computed using \textit{cv2.findContours()}, and a minimum-area rotated rectangle representing each vehicle was fitted to maximize point values using \textit{cv2.minAreaRect()}.

\begin{figure}[ht]
    \centering
    \includegraphics[width=0.77\linewidth]{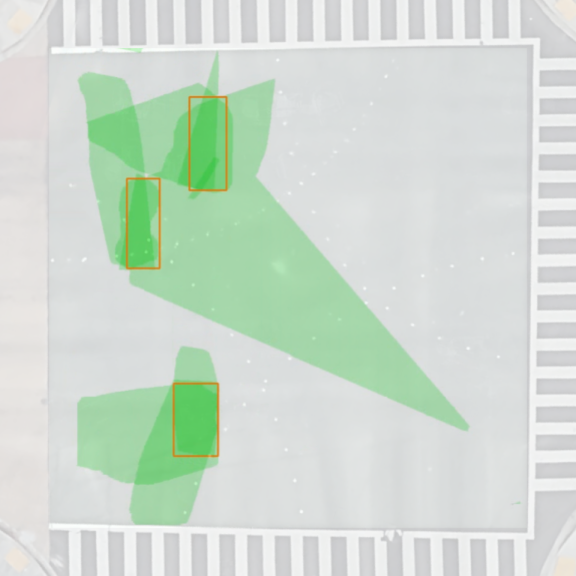}
    \caption{Rectangle fitting code operating on four layered camera views.}
    \label{fig:rectangle_fitting}
\end{figure}

Additional safeguards were applied to improve accuracy, including snapping rectangles to 0° or 90° when aspect ratio and area thresholds were met, extending rectangles near the intersection boundary based on inferred vehicle locations, and splitting excessively large rectangles along low-density midlines to separate adjacent vehicles. Variables and offsets were tuned based on testing data. Vehicles meeting coverage criteria were logged as rectangles with Unix millisecond timestamps in a MySQL database. A visualization of the rectangle fitting process is shown in Fig.~\ref{fig:rectangle_fitting}.

Table~\ref{tab:vehicle_rectangles} presents a representative sample of vehicle rectangle data recorded in the SQL database following the post-processing on the Windows system. Each entry corresponds to a single detected vehicle and includes coordinates, dimensions, orientation, and area. This dataset serves as the input for the PET calculation algorithm, which iteratively processes these entries to generate the corresponding traffic safety heatmaps.


\begin{table}[ht]
\centering
\caption{Sample recorded vehicle rectangle data}
\setlength{\tabcolsep}{2pt}
\begin{tabular}{|c|c|c|c|c|c|c|c|c|c|c|c|c|c|}
\hline
\textbf{Time} & \textbf{Length} & \textbf{Width} & \textbf{Angle} & \textbf{Area} & \textbf{$x_1$} & \textbf{$y_1$} & \textbf{$x_2$} & \textbf{$y_2$} & \textbf{$x_3$} & \textbf{$y_3$} & \textbf{$x_4$} & \textbf{$y_4$} \\
\hline
0.0 & 61.21 & 39.33 & 20.56 & 2407 & 200 & 187 & 214 & 0 & 272 & 172 & 258 & 208 \\
0.1 & 150.23 & 57.61 & 0.0 & 8653 & 205 & 238 & 205 & 88 & 263 & 88 & 263 & 238 \\
0.2 & 159.71 & 66.34 & 0.0 & 10595 & 205 & 281 & 205 & 122 & 271 & 122 & 271 & 281 \\
0.3 & 162.38 & 68.87 & 0.0 & 11182 & 202 & 320 & 202 & 157 & 271 & 157 & 271 & 320 \\
0.4 & 156.05 & 71.63 & 0.0 & 11178 & 200 & 353 & 200 & 197 & 272 & 197 & 272 & 353 \\
\hline
\end{tabular}
\label{tab:vehicle_rectangles}
\end{table}

\subsection{PET Calculation}

\begin{figure}[ht]
    \centering
    \includegraphics[width=0.65\linewidth]{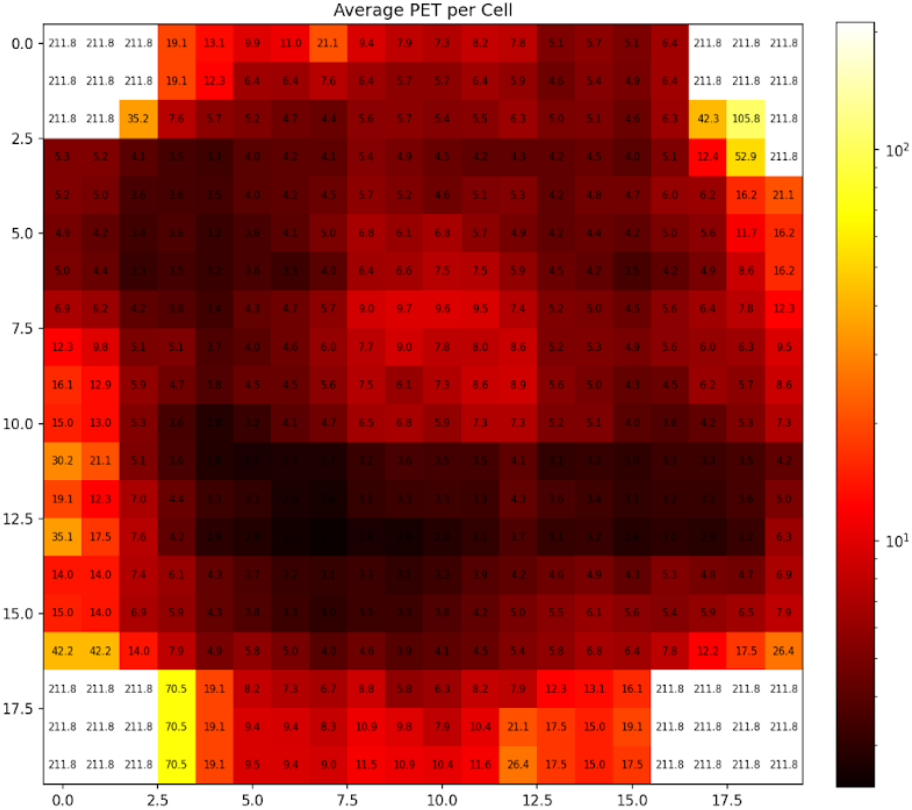}
    \caption{Cell-level PET heatmap over a 5-minute dataset for a $20 \times 20$ grid\newline red = lower PET (hazardous), white = higher PET (safe).}
    \label{fig:coarse_heatmap_1}
\end{figure}

Post-Encroachment Time (PET) was computed for each pixel in the intersection. Initial computations employed a $20 \times 20$ coarse grid, shown in Fig.~\ref{fig:coarse_heatmap_1} and the left side of Fig.~\ref{fig:json_blob_new}, reflecting common cell-based PET calculation techniques identified in existing literature. This was later extended to an unbounded resolution grid, allowing arbitrary granularity depending on the specified grid resolution. This approach enables a fine-grained visualization of vehicle movement and hazard locations without the obfuscation inherent in coarsely binned grids.

In the fine-grained visualization method, a “stopwatch” value $s_{i,j}(t)$ was maintained for each pixel $(i,j)$ at time step $t$. The update rule is defined as:
\[
s_{i,j}(t) =
\begin{cases}
0, & \text{if vehicle occupies $(i,j)$ at } t \\
s_{i,j}(t-1) + \Delta t, & \text{if pixel $(i,j)$ unoccupied at } t
\end{cases}
\]
\noindent where $\Delta t$ is the time interval between successive frames. Whenever a pixel transitions from occupied to unoccupied (i.e., $s_{i,j}(t-1) > 0$ and $s_{i,j}(t) = 0$), the previous value is logged and used to update a rolling maximum PET value $\overline{\text{PET}}_{i,j}$ for that pixel:

\[
\overline{\text{PET}}_{i,j} = \frac{1}{N_{i,j}} \sum_{k=1}^{N_{i,j}} \text{PET}_{i,j}^{(k)}
\]

\noindent where $N_{i,j}$ is the number of occupancy intervals recorded for pixel $(i,j)$, and $\text{PET}_{i,j}^{(k)}$ is the duration of the $k$-th unoccupied interval. The resulting values represent the average time each region of the intersection remains empty between vehicle passages, providing a continuous and high-resolution measure of traffic conflict risk.

The pixel-level PET analysis is implemented within a localized region of interest of $800\times800$ pixels, centered in a larger $\begin{matrix}1600\times1600\end{matrix}$ grid to cover approximately $26.2 \times 26.2$ meters of the intersection. This provides a spatial resolution of roughly $3.3 \times 3.3$ centimeters per pixel. To filter out noise and momentary occupancy, a time threshold is applied where a pixel must remain unoccupied for at least 350 milliseconds to contribute to a PET interval. This was chosen so that a pixel should be unoccupied for at least the duration of one fully processed frame. Post-encroachment intervals are averaged over the dataset duration to generate the final mean PET value per pixel. To complement this temporal data, a linear update count is also maintained for each pixel to track the total number of PET events, providing a secondary measure of traffic density patterns alongside the safety conflict metrics.

\subsection{Data Storage}

\begin{figure}[ht]
    \centering
    \includegraphics[width=1.0\linewidth]{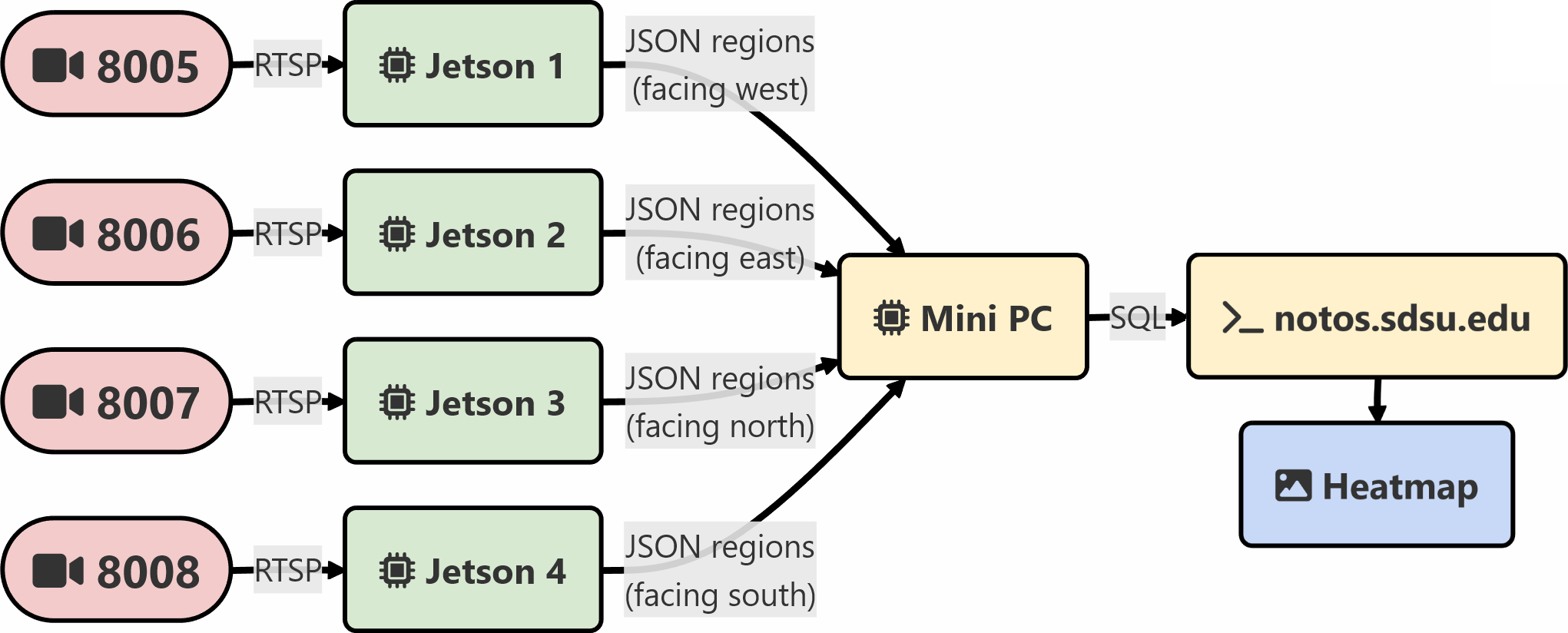}
    \caption{Overall data pipeline from cameras to heatmap.}
    \label{fig:data_flowchart}
\end{figure}

Each of the four NVIDIA Jetson AGX Xavier devices was paired with one PTZ camera and was responsible for processing the video from that camera independently. Following image segmentation and coordinate conversion on a given Jetson device, the resulting JSON files were named according to their Unix millisecond timestamp and saved to a shared folder accessible by both the Jetson device and the Windows system via the Server Message Block (SMB) protocol. On the Windows system, a Python script monitored the four camera-specific folders and identified sets of JSON files with the closest timestamps across all cameras. This structure is visible in Fig.~\ref{fig:data_flowchart}.

The script retained the timestamps of previously processed JSON files as temporary variables to prevent re-selection in subsequent frames. A maximum allowable difference of 350 milliseconds between the earliest and latest timestamps was enforced. The average of the four timestamps, rounded to the nearest millisecond, was recorded when logging the coordinates of the fitted rectangles in the MySQL database. If a complete set of JSON files from all four cameras within the 350-millisecond tolerance was unavailable, rectangle fitting was attempted using only three cameras. If a three-camera set also failed to meet the tolerance, the frame was skipped. During testing, the processing time per frame was approximately 350 milliseconds, which defined the maximum tolerance for timestamp alignment. Exceeding this threshold risked incorrectly merging frames from different time points despite the availability of closer-synchronized data.

\begin{figure}[ht]
    \centering
    \includegraphics[width=1.0\linewidth]{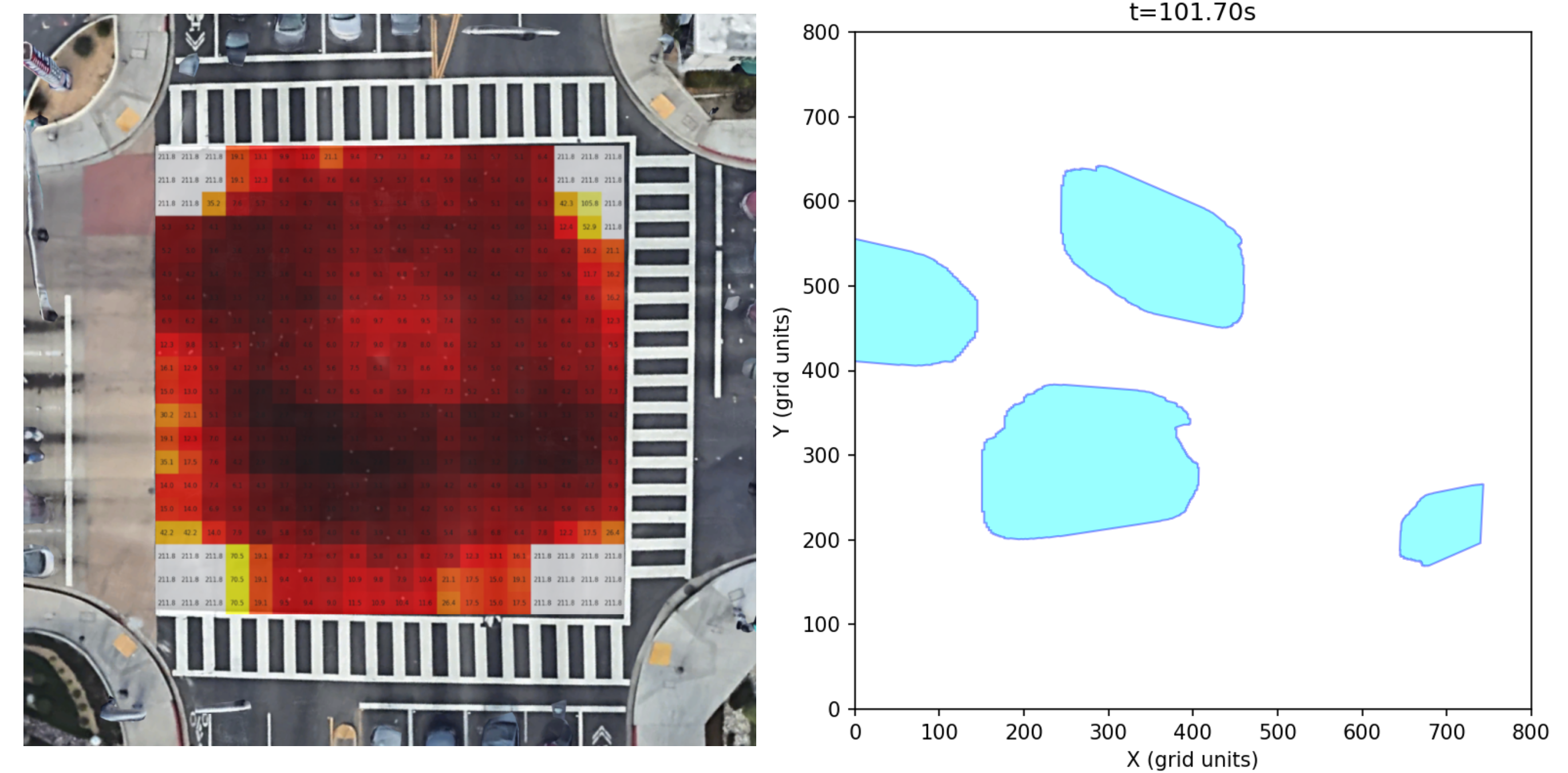}
    \caption{LEFT: PET heatmap on intersection, Red = lower PET (hazardous);
    \newline
    RIGHT: Visualization of information stored in JSON files. Each contour represents the predicted vehicle location from one camera's perspective.}
    \label{fig:json_blob_new}
\end{figure}

Both the original JSON files and the processed rectangle coordinates stored in the MySQL database were retained, enabling the creation of an open-ended dataset. This dataset can be re-indexed or reprocessed at any future time using improved rectangle fitting or PET calculation methods. An example visualization of a single JSON file is presented in the right panel of Fig.~\ref{fig:json_blob_new}.

\subsection{Implementation}

To test the efficacy of the model running locally on the Jetson devices, diagnostic data points such as core usage or rectangle creation speed were logged for different methods.

As the locations of rectangles were saved to the SQL database and deleted from the Windows machine, a separate piece of Python code designed to generate PET heatmaps was run on the \textit{notos.sdsu.edu} GPU server for faster computation. \textit{notos.sdsu.edu} is a Supermicro SuperServer 4028GR-TR GPU cluster optimized for AI, deep learning, and/or HPC applications. \textit{notos.sdsu.edu} features 8 Nvidia Tesla V100-PCIe GPUs and has the TensorFlow, TensorFlow Lite, Caffe, and Keras deep learning frameworks installed.


\section{Results}

\subsection{Camera Coverage}

\begin{wrapfigure}{l}{0.2\textwidth}
    \includegraphics[width=0.2\textwidth]{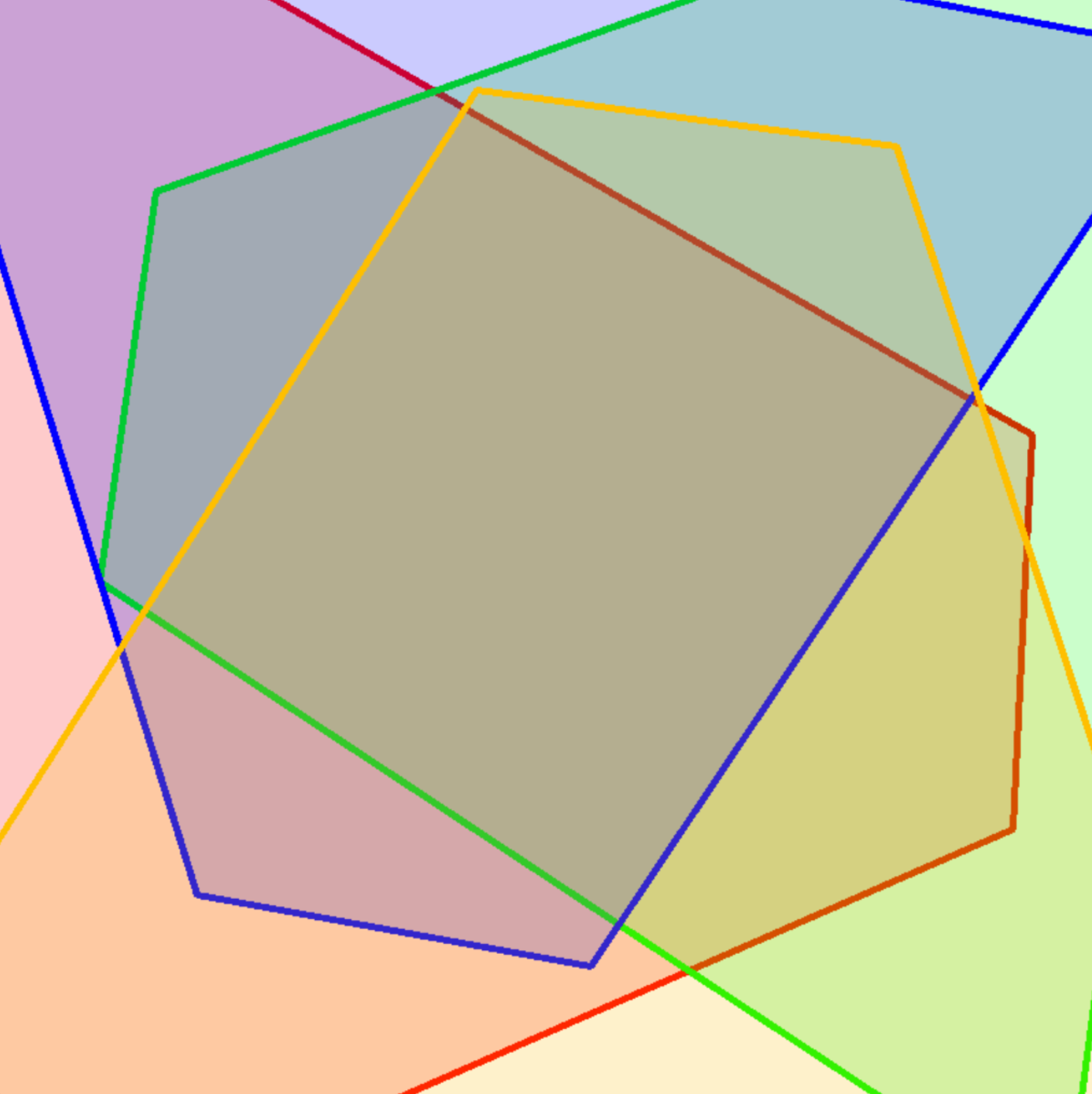}
    \caption{Camera overlap regions: Red = 8005, Yellow = 8008, Green = 8006, Blue = 8007.}
    \label{fig:overlaid_views}
\end{wrapfigure}

Once the homography matrices for all cameras were computed, the span of each camera’s field of view was projected into a common global coordinate system, as illustrated in Fig. \ref{fig:overlaid_views}. This projection revealed that the majority of the intersection was simultaneously observable by multiple cameras, providing critical redundancy for vehicle localization. 


The multi-camera coverage mitigates the inherent limitations of deriving 3-dimensional positions from 2-dimensional video frames, where occlusions and assumptions of ground-level vehicle positions can introduce errors. By leveraging overlapping viewpoints, the system reduces uncertainty associated with objects hidden behind vehicles, ensuring more accurate mapping of traffic participants across the intersection.

\subsection{Performance Metrics}

System profiling was conducted to evaluate the computational efficiency and real-time feasibility of the multi-camera vehicle detection and PET computation pipeline. Across all four NVIDIA Jetson AGX Xavier devices, the average total processing time per frame was 372.7 ms, corresponding to an effective frame rate of 2.68 FPS. The image segmentation inference using YOLOv11 and coordinate conversion accounted for approximately 77.5\% of the total processing time, averaging 288.8 ms per frame. Data transmission from the cameras contributed 83.8 ms per frame, mostly due to occasional processing errors or RTSP decoding errors.

On the Windows system, the rectangle fitting algorithm exhibited an average processing time of 20.3 ms per frame, with SQL database uploads averaging 32.5 ms, though this happened concurrently with the image segmentation and coordinate conversion code running on each Jetson device, so this did not affect the final frame rate. The PET value update process on the Notos computation node completed in 126.4 ms per frame, demonstrating efficient temporal aggregation performance.

Averaging values over a five-minute sample, overall CPU utilization during peak load was approximately 41.7\% on the Jetson devices (10.4\% average per core) and 86.6\% on the Windows system, while GPU utilization on the Jetson devices averaged 52.6\%. The pipeline achieved near real-time performance at 2.68 FPS, even after RTSP decoding errors and processing latency, confirming that it can sustain continuous multi-camera processing. Profiling further indicated that the most computationally intensive components were the image segmentation inference and coordinate projection stages, which was the bottleneck that needs to be tackled in future work.

\subsection{PET Analysis and Findings}

\begin{figure}[ht]
    \centering
    \includegraphics[width=1.0\linewidth]{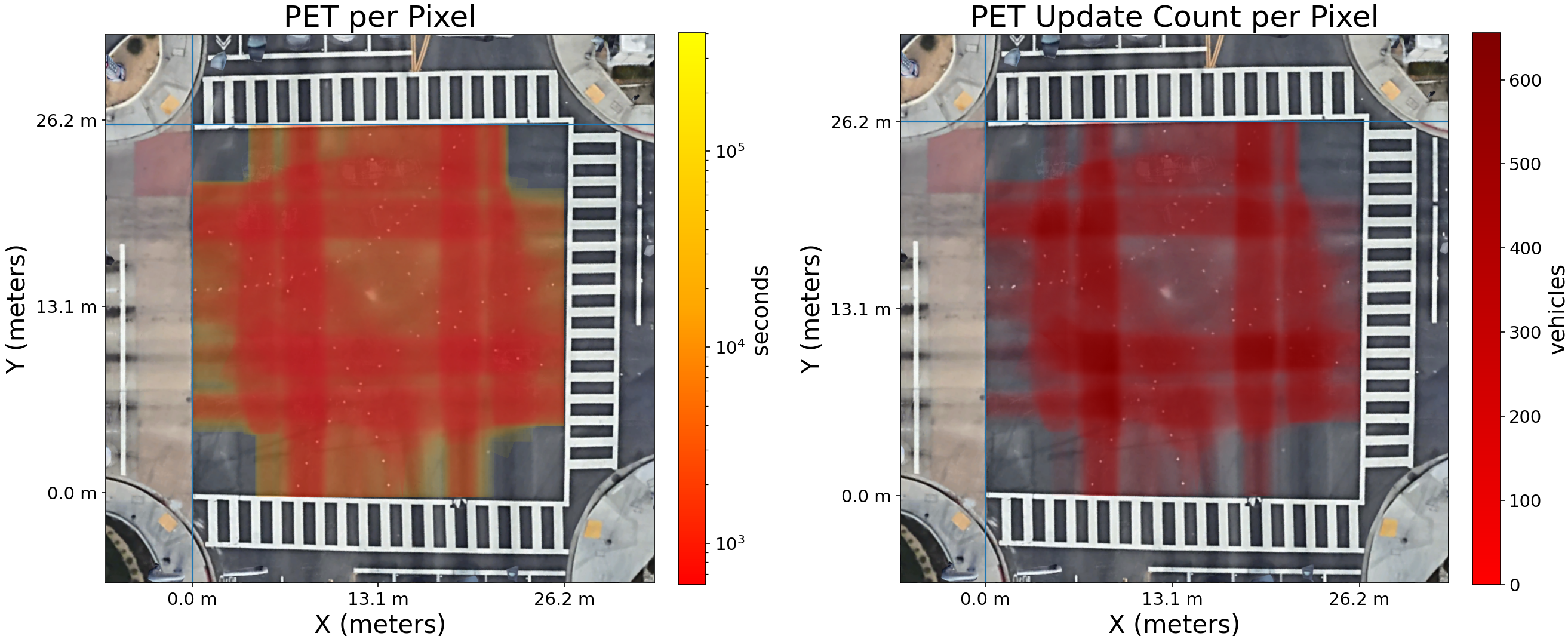}
    \caption{High-resolution pixel-level PET heatmap on 70-minute dataset\newline LEFT: Yellow = Higher PET (safe), Red = Lower PET (hazardous)\newline RIGHT: Red = more vehicles passed, White = fewer vehicles passed}
    \label{fig:fine_heatmap_1}
\end{figure}

The resulting high-resolution PET heatmaps in Fig. \ref{fig:fine_heatmap_1} reveal distinct patterns of traffic conflict across the intersection. By applying a logarithmic colormap, the regions with low PET values are successfully highlighted, which denote areas of frequent vehicle interaction and elevated hazard. The update count heatmap further confirms these findings by correlating high-risk zones with dense traffic patterns. Specifically, the pixel-level granularity identifies high-risk conflict zones that would be obscured in coarser analyses, providing actionable data for traffic management and potential intersection design improvements.

Analysis of the PET heatmaps revealed previously unrecognized hazardous movement patterns. In particular, northbound vehicles executing left turns exhibit relatively short PET values, indicating elevated risk in these maneuvers. Additionally, the lanes traveling straight immediately adjacent to the left turn lane, as well as the second lane over from the left turn lane, demonstrated the lowest PET values, indicating vehicles moving in close proximity to one another while proceeding straight, particularly after the traffic light turns green.

\section{Expansion}

\subsection{System Limitations and Future Improvements}

Vehicle rectangle fitting accuracy was limited by insufficient camera overlap, as at least three cameras were required to generate world-coordinate bounding boxes. This created blind spots along edges and corners of the intersection, preventing PET-based hazard calculations in these regions. Future implementations could adopt alternative indexing strategies to better utilize partially observed areas.

System generalization to new intersections is also constrained by hard-coded parameters. Coordinate conversion relies on manually annotated points of interest, and homography matrices were computed without correcting for fisheye distortion or camera-specific intrinsics. Reducing manual configuration and improving calibration would streamline deployment.

Vehicle tracking is another area for improvement. Currently, each detected vehicle is logged independently per frame, without temporal continuity. Incorporating multi-object tracking methods, such as DeepSORT, would enable consistent IDs across frames, off-screen vehicle recovery, velocity estimation, vehicle type aggregation, and trajectory prediction.

Edge device deployment introduced performance limitations. CPU usage and communication latency between the Jetson devices and the Windows system limited frame rates. Centralizing computation on a more powerful machine would reduce delays and support additional processing. As PET calculations require synchronized inputs from all four cameras, uneven offloading is impractical, making centralized processing the most scalable approach.

\subsection{Web Interface}

The JSON and rectangle datasets cover a wide range of timestamps, enabling PET calculations for specific intervals, such as times of day or days of the week. A web interface could allow users to generate PET heatmaps for custom intervals, supporting real-time hazard assessment and improving system utility.

\subsection{Road Design Recommendations and Significance}

Adjusting yellow light durations and strategically deploying hazard lights could help reduce high-speed conflicts by providing drivers with clearer warnings and reaction time at critical areas of the intersection.

The PET heatmap data reflect high traffic hazard at locations where lanes traveling straight are close to each other. To reduce the potential for high-speed accidents, the data suggest that more spacing should be added between straight-traveling lanes, possibly by widening the lanes. It appears from the data that left turns do not cause as much hazard due to their reduced aggregate vehicle density and vehicle velocity.

The significance of this work is that it demonstrates a practical and scalable method for generating continuous, pixel-level PET measurements using only multi-camera video data. This provides cities with a low-cost approach to evaluating traffic safety without requiring roadside sensors or connected-vehicle infrastructure. The resulting PET heatmaps capture fine spatial patterns of vehicle conflict that are not visible through traditional crash data, allowing hazardous areas to be identified before collisions occur. The modular pipeline also creates a foundation for future capabilities such as vehicle trajectory analysis, adaptive signal timing, data-informed road design, and the installation of devices to reduce pedestrian-vehicle conflicts, such as rectangular rapid flashing beacons (RRFBs), that activate when relatively low PET values are observed. This system offers a useful tool for ongoing and proactive safety monitoring.

\section*{Acknowledgment}

This research was supported by the Office of the Assistant Secretary for Research and Technology, Department of Transportation, Project Number: 04-110,
\textit{Developing an Intelligent Transportation Management Center (ITMC) with a Safety Evaluation Focus for Smart Cities}.





\printbibliography

\end{document}